# Deep-Learning Quantitative Structural Characterization in Additive Manufacturing

**Authors:** Amra Peles [1, *], Vincent C. Paquit [2], Ryan R. Dehoff [3]

**Affiliations:**

[1] Computational Sciences and Engineering Division, Oak Ridge National Laboratory, Oak Ridge, TN, USA

[2] Electrification and Energy Infrastructure Division, Oak Ridge National Laboratory, Oak Ridge, TN, USA

[3] Manufacturing Science Division, Oak Ridge National Laboratory, Oak Ridge, TN, USA

[*] Corresponding Author. Email: pelesa@ornl.gov

**Abstract:**

With a goal of accelerating fabrication of additively manufactured components with precise microstructures, we developed a method for structural characterization of key features in additively manufactured materials and parts. The method utilizes deep learning based on an image-to-image translation conditional Generative Adversarial Neural Network architecture and enables fast and incrementally more accurate predictions of the prevalent geometric features, including melt pool boundaries and printing induced defects visible in etched optical images. These structural details are heterogeneous in nature. Our method specifies the microstructure state of an additive built via statistical distribution of structural details, based on an ensemble of collected images. Extensions of the method are proposed to address Artificial Intelligence implementation of developed machine learning model for in real time control of additive manufacturing.

**Keywords:** deep learning, melt pool, additive manufacturing, process structure property relations, microstructure prediction

## 1. Introduction

***Additive manufacturing and importance of melt pools for quality and process improvement:***

Additive manufacturing (AM) is a fast-developing fabrication method aimed to produce fit-for-purpose components by design [1]. Unlike conventional subtractive manufacturing, it has significant advantages, such as a high degree of freedom in geometric design. It also promises simultaneous control of component geometry and materials properties at local component level. In the past few decades, AM technology has developed rapidly and has had a significant impact on the manufacturing industry opening new frontiers for engineering applications in various industrial sectors [2, 3]. The ability to print complex geometries and the potential to overcome deficiencies of current manufacturing processes makes metal additive manufacturing the most researched and fastest growing area of AM [3, 4]. Laser powder bed fusion (LPBF) is the most prevalent method for manufacturing of metal components with a near-net shape defined by a digital model. In this method, material in powder form is spread over a build plate in the layer of controlled thickness. The powder is then spot-melted by heat imparted by a laser. The melted spot subsequently cools down and solidifies as the laser moves along a predefined raster path over a deposited layer of powder [5]. During this process, a large number of complex heat and mass transport phenomena is taking place, yielding distinct melt pool geometries visible in the microstructure of manufactured components [6]. Correlations between the quality of manufactured parts and their multi-scale sub-millimeter structural features is well known. Besides, structural features are dictated by the manufacturing parameters [7, 8]. There have been a number of efforts linking processing parameters with melt pool geometry [9-12]. Understanding and utilizing these correlations is at the core of AM technological advancement and requires accurate, fast, and reliable ways to identify and quantify structural features themselves. One of the main challenges in the LPBF process is making dense and defect-free components. It has been shown that lack of fusion and keyhole porosity defects are dependent upon the melt pool geometry and the processing conditions [13]. Indeed, manufactured part-to-part variability in quality and presence of structural defects that result in performance failure present a barrier to the broader adoption of additive manufacturing, especially in industries like aerospace, automotive, and medical devices where human safety is of paramount importance [14]. Overcoming these difficulties requires fast and robust characterization of melt pool geometries to enable their correlations to both manufacturing parameters and part quality. However, finding fast, accurate, and efficient methods to spatially isolate and quantify melt pool geometries in structural images presents a challenge [6]. Inspection of melt pool features in additively made structures over a variety of materials reveals structural heterogeneity and stochastic characteristics. Consequently, their analysis needs to include a statistically relevant quantity of data to extract information relevant to material behavior. While the human eye is well versed in spotting and labeling the melt pool features in an image, it is an unsustainable way to obtain necessary statistical accuracy, hence, systematic algorithmic approaches are required. Recent advances in computer vision and machine learning suggest their use in robust structural characterization of images of AM components should be explored. Machine learning as a technology development accelerator has been shown in a wide range of applications in recent years fueled by the large amounts of data being generated in AM [15, 16]. However, data set completeness, data variety, veracity, and validity remain major challenges for construction of well performing machine learning models on such data. Models' pertinence, transferability and interpretability need to be better understood. Reported machine learning studies focused mainly on predicting melt pool widths and heights from processing parameters based on available data sets either collected from literature [17], or created for the purpose of understanding processing-structure-property relationships [6, 18, 19, 20]. Instead, we develop a method and a set of tools for melt pool boundaries prediction directly from structural images and subsequent statistical analysis of the melt pools' geometric features. Our approach uses an image-to-image translation generative adversarial neural network (GAN) for deep learning of melt pools and defects structural features from optical images of AM components. Reported capability is promising for accelerating AM parts qualification and materials process certification, as well as fundamental process improvements, optimization, and control, thus fostering faster adoption of AM. In addition, our method promises to overcome key barriers for in-situ monitoring and to lead towards development of artificial intelligence AM process control.

*Deep Learning Methods:* Larger amounts of available data allowing for more artificial neural networks layers, and introduction of the concept of convolutions have made deep convolutional neural networks (CNN) very promising for technology applications. In the area of understanding and quantifying additively manufactured microstructures, several methods have been tested. An analysis of thermal imaging data to improve image contrast ratio of a melt pool thermal images have shown that adding a constraint term to a plain GAN loss function yields a better contrasting image [30]. In the additive process monitoring study, CNNs were used to classify melt pools based on their size when analyzing images of evolving melt pools created by a moving laser beam [18]. Semantic segmentation of melt pool images on a U-net CNN architecture were reported on augmented data set annotated and post processed using the water-shading algorithm [31]. The GAN, an unsupervised or semi-supervised deep learning model, has facilitated a significant breakthrough in a number of domain areas [21]. A GAN could serve as a generative model that aims to produce samples of data that are statistically equivalent to those from real data set. In this context, the advantage of GANs is their ability to be self-trained with a minimal amount of data. Indeed, the statistical nature of melt pool distributions confirms the rationale for exploiting the GAN approach as a generative model, which is trained on annotated melt pool boundary data in a self-consistent way with improved accuracy at each step. In general, GAN models consist of two deep learning networks referred to as generative and discriminative models. The generator neural network extracts the hidden regularities in input data and generates output sample data based on the parameters from extracted underlaying regularities. These samples, together with real samples, serve as an input to the discriminative model, which is a binary classifier with an output probability. This probability represents the model confidence that generated input is possible in a real data set. Image to image translation GAN based generative modeling is promising for learning how to determine presence of melt pools and delineate its boundaries from extracted regularities [22]. Here we examine the effectiveness of the image-to-image translation GANs deep learning on predicting the melt pool boundaries and defects in the images and its possible applications toward overcoming key bottlenecks facing the qualification and process improvement in additive manufacturing. In addition, we introduce the workflow that enables higher accuracy and speed of delineating the melt pool boundaries with minimal annotation effort.

*Quantitative and statistical analysis of melt pool geometries:* The heterogeneous and stochastic nature of AM-built structure precludes a deterministic description. Usually, structure is described and quantified via an intuitive list of statistical measures such as size and shape of representative melt pools, defects, and their probability distribution [6]. Two-point statistics, similar to the pair correlation function structural description in ordered solids, was introduced to describe spatial correlations between distinct local phases in the structure of two-phase alloys. It was shown that given symmetry constraints, it was possible to reconstruct structures from these statistical data, demonstrating the importance of statistical insights in the structural features [23, 24]. A disadvantage of previous methods is the computational cost that practically precludes analysis beyond two-point statistics. In addition to predictive capability of localized delineation of melt pools and defects in microstructure images, we created a set of tools to extract intuitive statistical representations based on the available data. Carefully quantifying these key features from AM built microstructure images and its statistical measures is a necessary step in establishing and using processing–structure–property relations.  This paper presentation follows chronological order of research activities. The methods section starts with a description of data pre-processing and initial melt pool annotation strategy that was later replaced by a deep learning model. The details of deep learning and iterative workflow methods are presented next. The results section discusses deep learning model convergence, performance predictive ability with increasing size of annotated data set; followed by the quantitative analysis and statistics of melt pool geometric features. The conclusion and outlook for future efforts are presented at the end.

## 2. Methods
### 2.1. Data Set Description and Preprocessing of Images

An assimilated data set consists of optical images taken on a grid of points on a cross-sectional area of an additively built geometry. The build powder is a multicomponent Al alloy with rich microstructural features at sub-micron scale and typical melt pool features at higher length scale which was probed in this study [25]. To better reveal geometric features of melt pool boundaries, the cross-sectional area was chemically etched. The cross-section of the build geometry is shown in Figure 1a with marked grid of points where higher resolution images were taken. Each image is 1920 x 2560 pixels in size with one distinct color channel. Examples of optical images are shown in Figure 1b. Loop-like melt pool geometries and distinct processing defects are apparent to the human eye. For distinction, processing defect features on images were segmented in green in Figure 1c. Overall, there are ~300 independent images in the data set with statistically relevant diversity of available data.

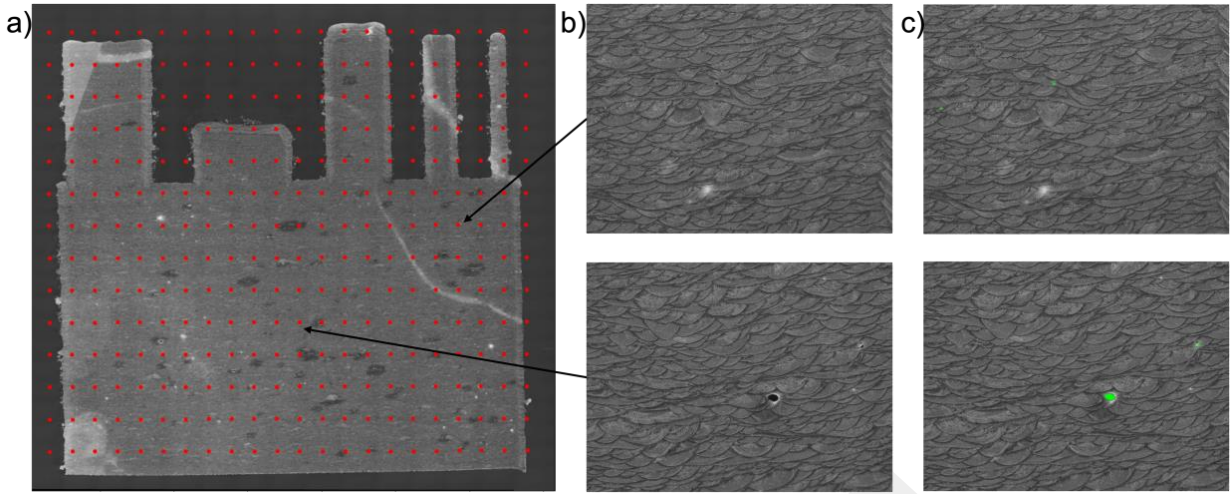

*Figure 1. a) cross-sectional area of a printed part with grid of points where higher resolution images where taken, b) original high resolution optical image samples showing visually distinct melt pools and processing defects, c) processing defects, in b) are segmented in green.*

The quality of etched specimen images is suitable for expert annotation of boundaries. In other words, an expert can visually distinguish neighboring melt pool features and identify the boundary between them. However algorithmic segmentation does not lead to fully closed boundaries around melt pools. This has been a known deficiency of conventional image processing methods [26]. By image processing segmentation, we mean known ways to determine for every pixel if it is part of the melt pool including the melt pool boundary. In particular, methods such as water-shading, thresholding, or edge detection fail to accurately predict the expert drawn boundary. Due to apparent bright-dark color contrast in images shown in Figure 1, the initial annotation to enable machine learning was done with the help of thresholding segmentation and expert input on definition of melt pool boundary. The melt pools segmented using thresholding are identified in white and depicted in Figure 2a. We use computer vision (OpenCV) tools to prepare initial training data set for a semi-supervised self-driven learning [27]. The conventional binary thresholding is used as a starting point with threshold value corresponding to the mean of pixel values. We also implemented Gaussian smoothing in conjunction with smearing filters of varying kernel sizes, which provided a reasonable initial binary segmentation, as depicted in Figure 2a.

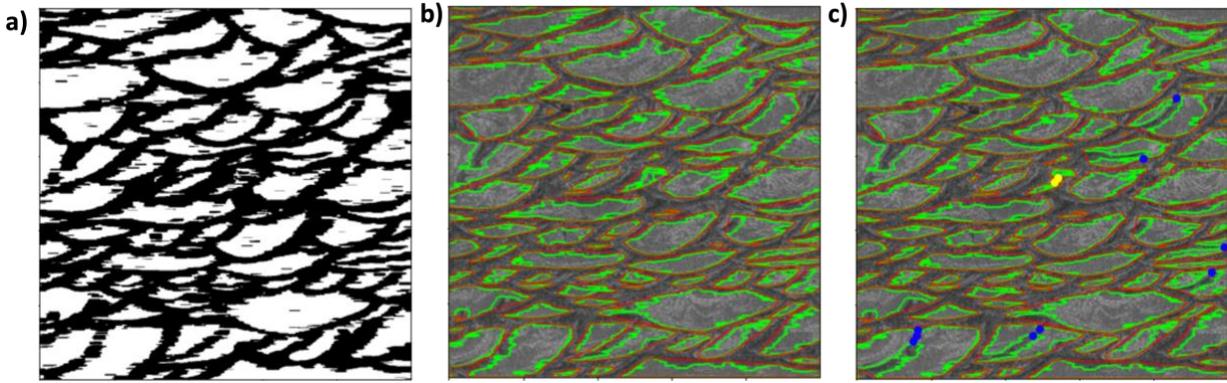

*Figure 2. a) binary thresholding, b) image with drawn boundaries per thresholding results c) thresholding flaws identified by expert input. Blue points denote where the thresholding boundary encompasses more than one melt pool. Yellow points denote where the thresholding dissects single melt pool into two or more.*

By drawing and inspecting binary segmentation contours on the image as shown in Figure 2b, we interactively identify points where the segmentation method failed. The manually inputted blue points identify lack of melt pool boundary, while yellow points identify segments that should form single melt pool. This input was then used to draw corrected melt pool boundaries algorithmically. Furthermore, the void-type processing flaws on images were also annotated. Figure 3a shows a result of our approach -- a 1020 x 2560 image with defects in green and melt pool boundaries in white. To create initial training data set, each annotated 1920 x 2560 image was divided into thirty-two 512 x 512 images. The division was done to minimize the image overlap and maximize variety of observed melt pool boundaries. Furthermore, 512 x 512 image excerpts from original images were compressed to a 256 x 256 size and used as inputs to model architecture shown in Figure 4. Data input to our deep learning model consists of image pairs, the raw data image, and corresponding annotated data image with melt pool boundaries and defects drawn in distinct color as illustrated in Figure 3b. The model is trained to take raw image input as those on the left in Figure 3b and to produce images on the right with melt pool boundaries drawn and defects identified. The initial data set had 192 data points. We describe deep learning model in the next section.

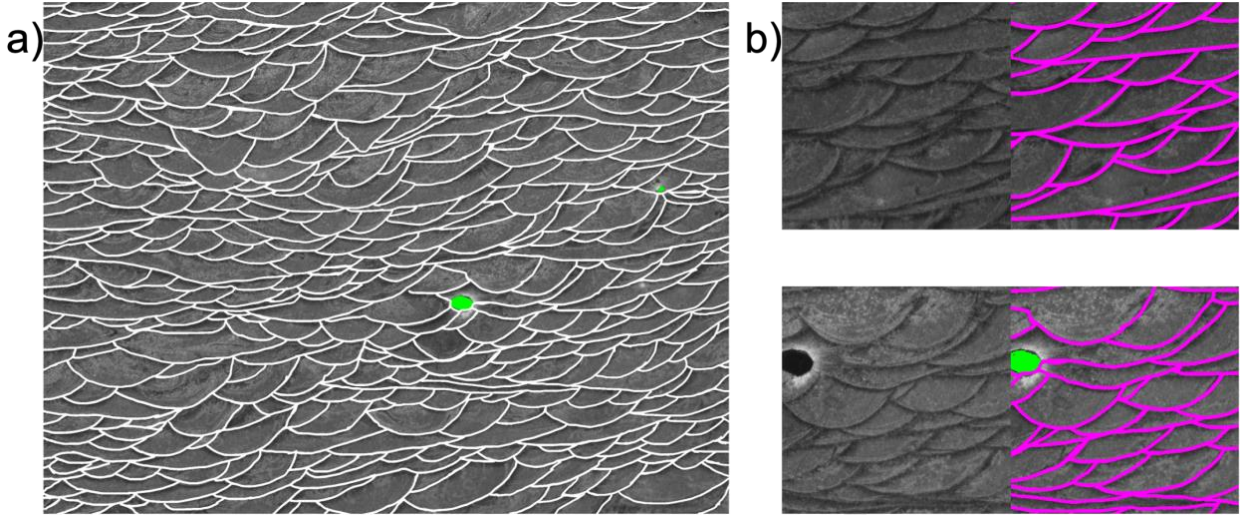

*Figure 3. a) annotated image for initial data set, b) model input examples with original image data on the left and annotated images with the melt pool boundaries (magenta) and defects (green) overlaid on gray background.*

## 2.2. Generative Adversarial Neural Networks for Image Translation

In our study, we adopt an image-to-image translation deep learning method with conditional generative adversarial neural networks [22]. This approach comprises a U-net generator [28] and a convolutional discriminator based on the similarity probability statistics of image patches (PatchGAN) [29]. As a result, each trained model translates images in the dataset from an original image into a target melt pool boundary annotated image. The model is trained to predict these annotations in a semi supervised manner. The model architecture is illustrated in Figure 4. The generator model G is conditioned to learn mapping G: $\{x, z\} \rightarrow y$, where x and y represent the observed original image and a model-learned annotated representation exercised on both original *x* and random noise *z*. This learning is supported by the discriminator model D $\{\{x, z\}, y\} \rightarrow p$, which outputs the probability of observed image y being generated by the G, thus improving the ability of G to create accurate annotations of melt pool boundaries.

The generator takes $256 \times 256 \times 3$ input with color channel values scaled to the [-1,1] range, and outputs images of the same size. The generator consists of encoder and decoder U-net wings. The encoder down-sampling has eight model blocks each lowering the size by half. First seven model blocks are each comprised of 4 x 4 kernel convolutions, a renormalization layer except for initial down sampling followed by the leaky rectified unit layer. A 4 x 4 kernel convolution in the last encoder step is followed by rectified unit activation. The decoder mirrors the encoder structure with eight upscaling steps. The first seven model blocks are comprised of 4 x 4 kernel transpose convolutions and batch normalization each, with incorporation of 50% dropout regularization in the first three blocks to introduce stochastic noise. Last step transpose convolution is followed by a hyperbolic tangent activation. Skip connections, illustrated by arrows in Figure 4, are made between layers of the same size in the encoder and decoder wing of the U-net architecture to mitigate down sampling/up sampling bottleneck.

The generator model learns to produce annotated images with the help of a discriminator model *D*. The discriminator is a convolutional "PatchGAN" classifier. The classification metrics are at the scale of image patches, evaluating the image similarity at local patch size level. The patch-wise evaluation predicts the likelihood of whether the observed image is from the annotated images data set or is created by the generator. The architecture of discriminator is shown on the right-hand side of Figure 4. Two input images are concatenated to a 256 x 256 x 6 input to the first convolution layer of the discriminator. It is followed by a series of convolutional (batch normalization) leaky ReLU blocks, the same as the generator down sampling structure, ending with "sigmoid" 16 x 16 x 1 final activation map. Each activation map value corresponds to a 70 x 70-pixel patch of the 256 x 256 input image and presents a probability in range [0, 1]. At the end, based on those probabilities, discriminator outputs indicate the model's confidence of observed image being the true annotation.

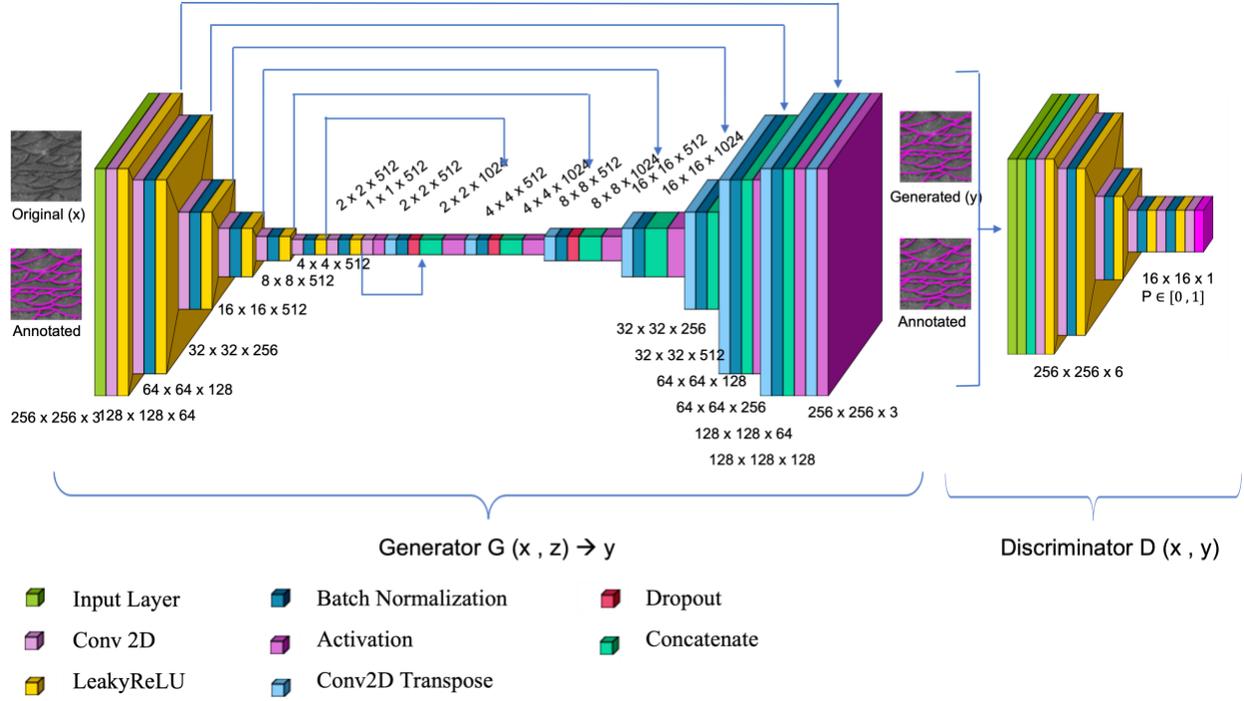

*Figure 4. Deep learning model architecture.*

The model optimization follows the loss function which incorporates an adversarial and image similarity losses. The loss function is expressed as:

$$\mathcal{L}_{c-GAN}(G,D) = \mathbb{E}_{x,y}[log(D(x,y)] + \mathbb{E}_{x,z}[log(1 - D(x,G(x,z))] + \lambda \mathbb{E}_{x,y,z}[\| y - G(x,z) \|_1].  \quad [1]$$

The adversarial nature of the model comes from competing objectives of two sub-models with respect to the loss function described in first two terms, namely, G tries to minimize it while at the same time D tries to maximize it. The third term compares the generated output to annotated images and forces the generator to remain close to the ground truth and produce outputs that are plausible translation of original images. The discriminator training is much faster than generator training, thus speed up of training is regulated via heuristic parameter $\lambda$, which is set to values between 80-200 in our study depending on the size of the input data set.

## 2.3. Iterative Annotation Workflow

The described model architecture is surprisingly efficient to learn and fit data even for data sets with sizes well below the number of model parameters. Initial fitted model predictions on the "unseen images" required significant expert input to fill in model predicted annotation gaps and correct eventual errors. We exploit these early successes to introduce a self-consistent workflow that repeatedly uses a model prediction from the present step to annotate new data and use that data to augment training data set for the next self-consistent repetition. As expected, the models were performing better at each repeated step as the data set size increased and accuracy of annotation improved. The workflow is depicted in Figure 5. After initial annotation and training of the model, the iterative workflow steps consist of: (a) current saved models evaluation and down selection of well performing models, (b) deploying this model to create image-to-image translation on new batch of unseen images, (c) review of model accuracy followed by corrections to the annotated data and the data addition to the existing training data set, and (d) retraining of the deep learning model on the augmented data set. At the end we follow with the quantitative and statistical analysis of collected data.

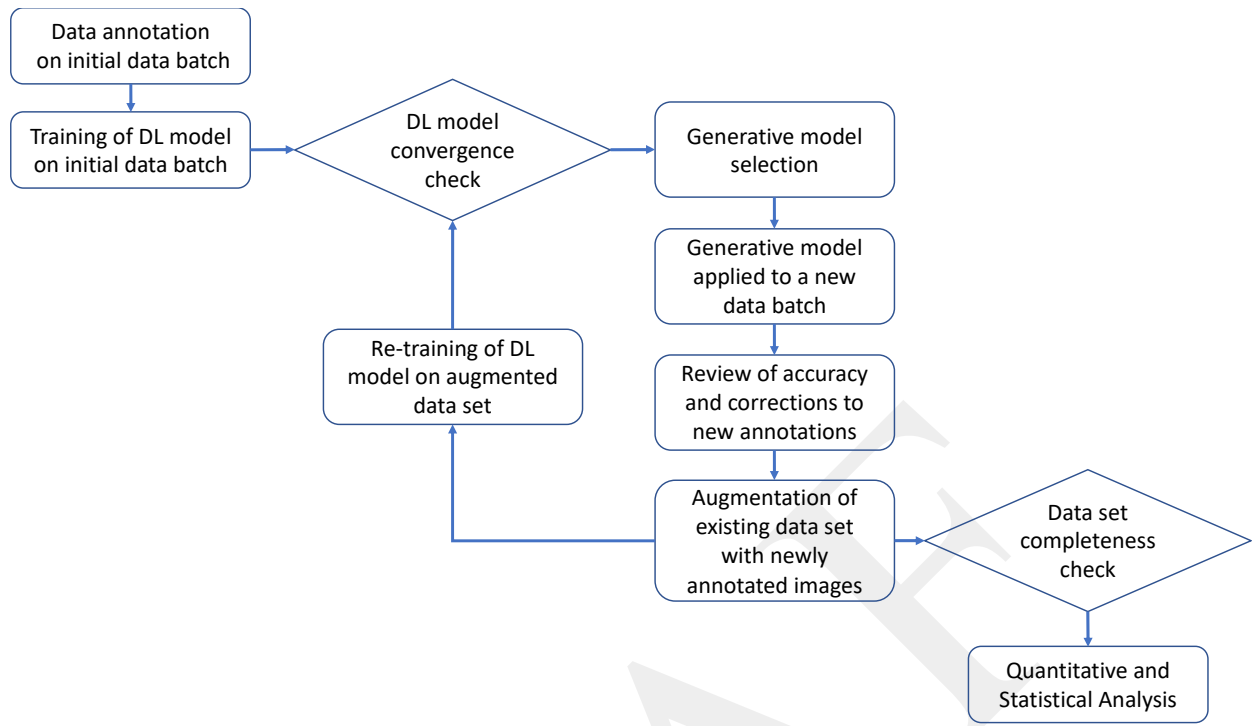

*Figure 5. The workflow for self-consistent improvements of model prediction and toward full automation of melt pool geometry boundaries determination and geometric features statistics.*

## 3. Results and Discussion

### 3.1. Model performance and training strategy

Starting with the annotated images of approximately 2% of the whole data set of optical images, we train the image translation GAN to learn image annotations and create a generative model. We implement generative model on the next batch of images from data set and create model predicted melt pool segmented images. Let us first discuss typical GAN-model performance by examining the changes of loss function (equation 1) with the number of training steps. The performance of the discriminator on real and generated data is shown in blue and orange in Figure 6, while generator loss is shown in green. The spikes in the loss data during training are due to simultaneous optimization of three competing losses and observed during a training on data divided in batches. The achieved generator loss errors went down to ~0.5 without observing significant overfitting in data. Overall, model convergence on our data was good. Due to the size of saved model files, we recorded generator model in predefined intervals of a total number of training steps, shown in red dots in the Figure 6.

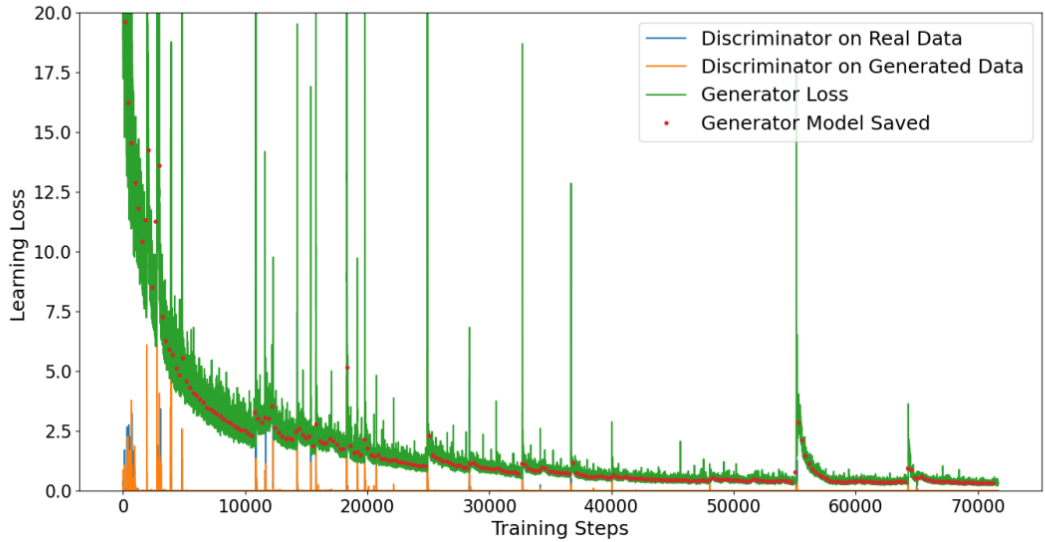

*Figure 6. The model convergence of Loss function with number of training steps.*

Model performance accuracy with respect to the ability to identify defects and melt pool boundaries depends on size of the input data set, accuracy of annotations, the number of training steps, and the loss function convergence. We evaluate an image structure similarity measure to gauge this performance and make data selection for new training data in our workflow. The chosen structure similarity approach assesses luminance, contrast and structure when comparing images, and quantifies image similarity by an index with values ranging between 0 and 1, where 1 means a perfect match between the compared images [33]. We use scikit image library to implement this method and obtain a structural similarity (SSIM) index and a difference image. Difference images help us determine where exactly the image differences are in terms of image coordinate location. Difference images are darkest where images differ the most. The white areas in difference images indicates identical parts of predicted and expected annotations. Next, in Figure 7 we show generative models' performances on the unseen images chosen randomly from the validation data set. Each row in the figure corresponds to a different generative model obtained from different size of the input data set -- 512, 768, 3200, and 7680 data points top-down, respectively. The columns in Figure 7 show respective unseen images in a), generative model predictions in b), expert annotated melt pool boundaries and defects ground truth in c), and difference measure between predicted and expected image in d). The top row in Figure 7, corresponds to a generative model trained on 512 data points set. This generative model could not predict the presence of defects and has an image similarity index of 0.85. The results for an increased input data set size to 768 data points are shown in the second row of Figure 7. The model predicts the presence of defect (in green) after convergence but shows signs of data overfitting, evidenced by the noise in predicted image. The noise distribution is also seen in the difference image. Presence of noise due to overfitting has lowered the similarity index for this larger data set to 0.75. Further increase in input data set to 3200 data points result in an increased similarity index of 0.90, while some noise is seen in the difference image. The last row shows the performance of generative model trained on the 7680 data points. The SSIM index for this model is 0.93 and difference image shows there is virtually no noise, hence indicating a more optimal model. It is important to note that the number of predictions of this model with significant errors was much less than in the smaller data set models. These results indicate the discriminator's ability to capture key features in the original images and interpret them as a melt pool boundary and defects. The model accuracy increases with the data set size, as expected.

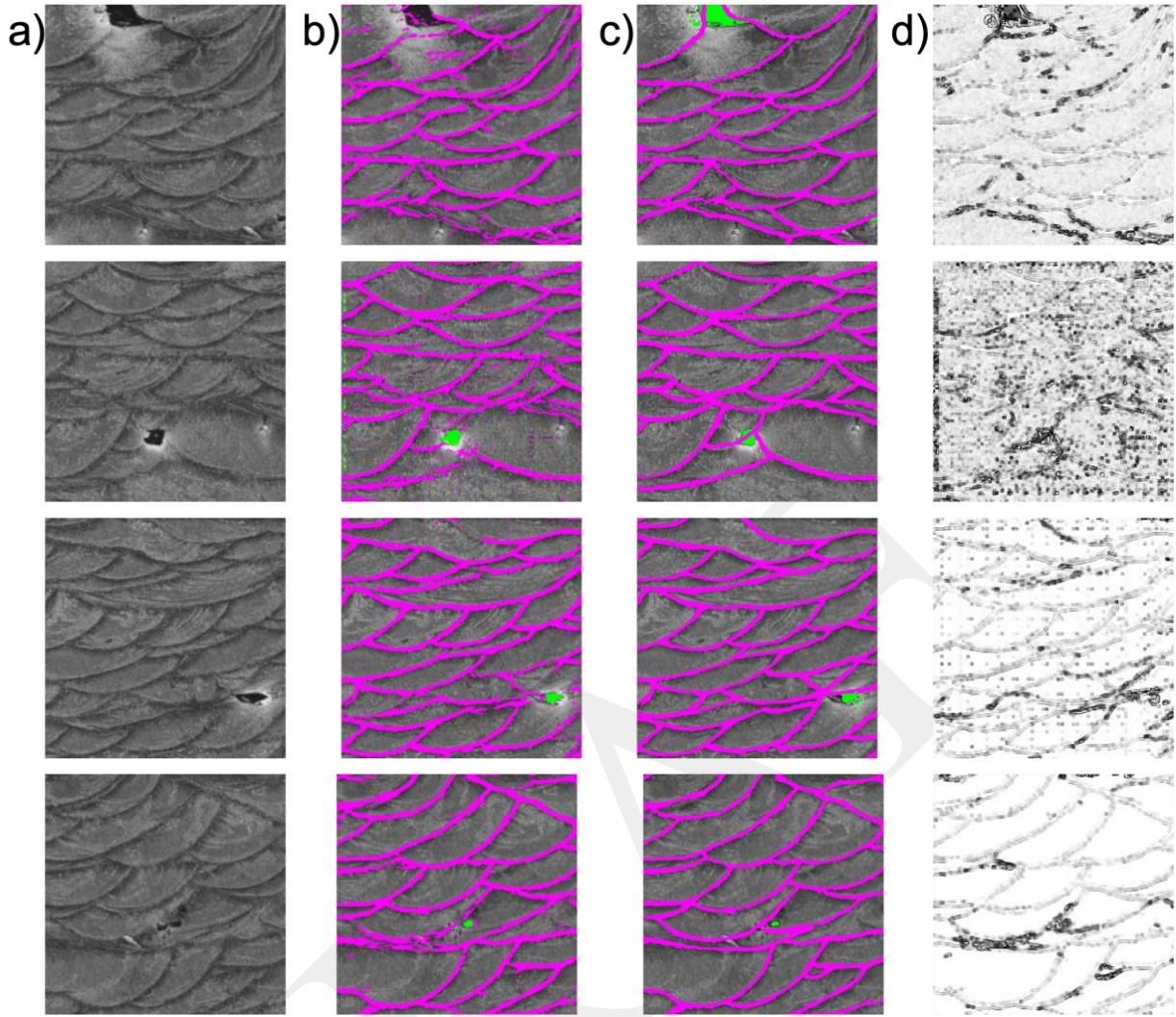

*Figure 7. a) validation data, b) models predicted melt pool boundaries and defects, c) expected annotations, d) difference image between predicted and expected images. The rows present data for models trained on 512, 768, 3200 and 7680 data set points with structural similarity index values of 0.85, 0.75, 0.9 and 0.93 respectively.*

## 3.2. Quantitative analysis and correlations of features

With good predictability of delineation between melt pools, we proceed with computer vision methods to extract melt pools geometric features and their statistical properties. Figure 8 shows the result of post processing of model image predictions using the code we developed. In a) we show melt pool boundaries, and in b) visible melt pool cross-sectional area differentiated by color for each melt pool. For single-track melts, melt pool width and height are commonly used geometric metrics because they are easily accessible from images since there is no effects of multilayer and neighboring tracks heating and mass transport. The insert in the Figure 8c illustrates the widely adopted single-track cross section geometry description. However, our data originates from the cross sections of fully built geometry and because of that shapes of identified melt pools have been affected by remelting during the processing. Remelting occurs both when the neighboring scan path is melted and also during the melting of new layers on the top surface of the part. Furthermore, laser scanning strategy typically used in AM maximizes variety of melt pool orientations by changing the laser angle by 67 degrees for each new layer. For this reason, a variety of melt pool shapes can be seen in images with different sizes and shapes. Inspired by thermomechanical model for the heat flux with a double-ellipse geometry, we implement ellipse fit to the melt pool boundary data to quantify melt pools properties [32]. The ellipse fit was constrained by the position of ellipse center and to bottom, not remelted portion, of the melt pool boundary.

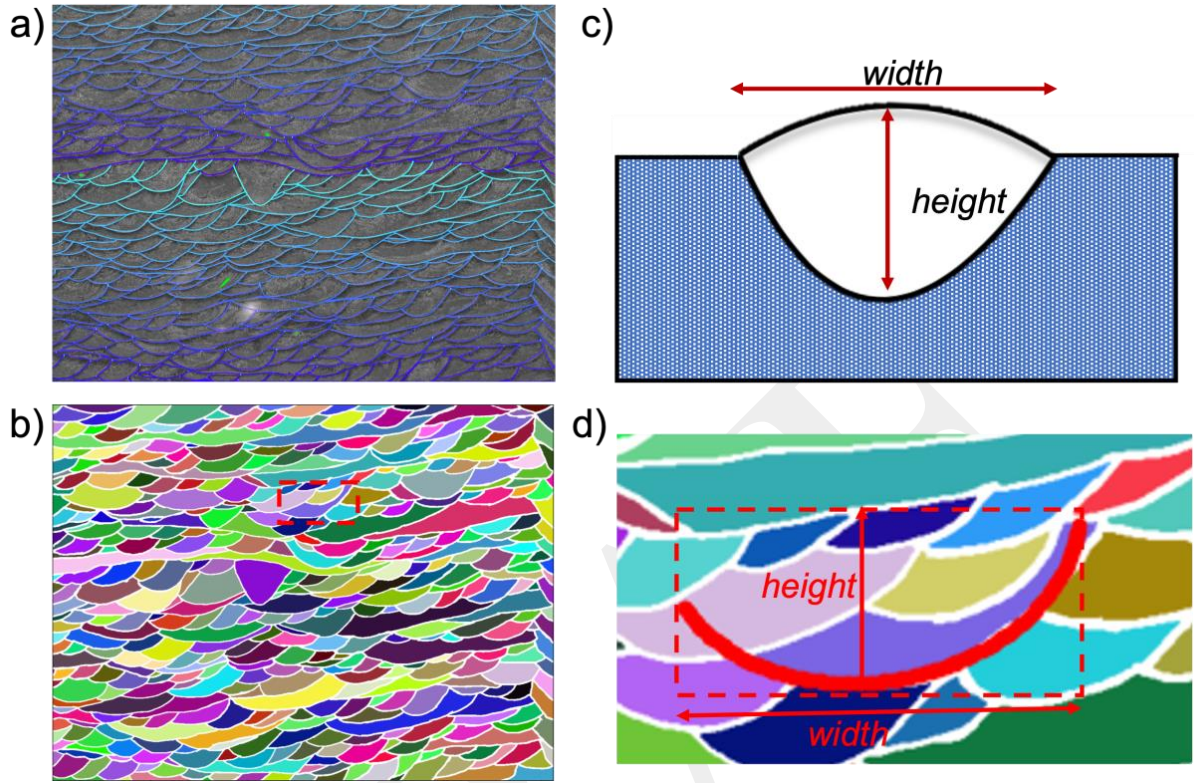

*Figure 8 Details of image processing: a) the post processed images of GAN model predicted segmentation of melt pools boundaries; b) Individual melt pools boundaries visible areas; c) schematic of a single-track melt pool geometry; and d) zoom in in marked area in b) showing arbitrary melt pool shape, elliptical fit to melt pool shape and definition of width and height in our study.*

Figure 8d shows enlarged boxed area from 8b together with fitted elliptical shape and how proxy melt pool dimensions are determined. This strategy allows for estimate of proxies of melt pool dimensions beyond visible bounds. Fit parameters were used to estimate apparent area as a half of the ellipse area and apparent aspect ratio as the ratio of shorter to larger half-ellipse dimensions. The melt pool fractions visible at the edges of images were not counted.

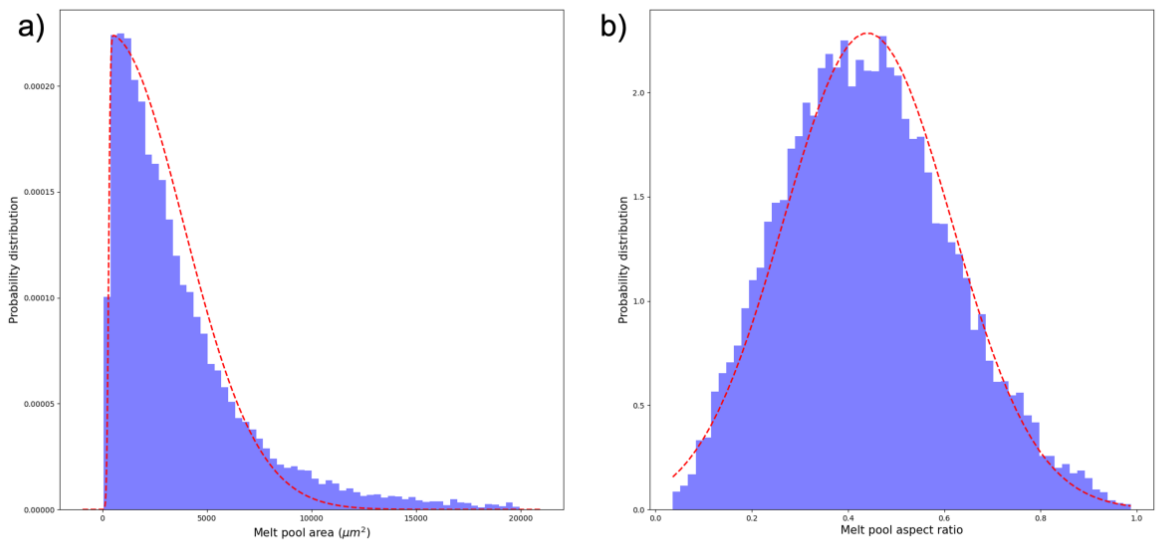

*Figure 9. The distribution of estimated apparent melt pool area and apparent aspect ratio of melt pool dimensions. Fits include skewness from normal distribution.*

Figure 9 shows the statistical distribution of apparent melt pool areas and apparent aspect ratios. The apparent area probability distribution is highly positively skewed toward higher values but dominated by the values peaking at about 400-500 micrometers. The apparent aspect ratio probability distribution has normal-distribution shape dominated by melt pools with apparent widths and apparent heights ratio of approximately 0.5. The results seem consistent with empirical expectations.

## 4. Conclusions and Outlook

We have demonstrated a fast and efficient pathway to structural characterization of additively manufactured parts. Our results show it is feasible to train image-to-image translation GAN to learn how to identify melt pool boundaries and defects in etched optical microscopy cross-sectional images. While our results are obtained for an optical images data set, the method can be easily generalized to other microstructural characterization images. The implemented workflow allows for integration of expert input and extended training of GAN models for improved accuracy. This presents a dramatic reduction in turnaround time compared to manual characterization of heterogeneous structural features. The methodology is also likely to be applicable in real time control of additive manufacturing parameters. Furthermore, our approach provides a set of tools to pre-process data for GAN model input and post-process data to extract melt pool and defect geometries and analyze them in statistically relevant quantities.

As the next steps, validation of predictive model transferability and transfer learning will be examined across different data modalities. The data modalities include builds with different materials, additive builds with different processing parameters, and variety of acquisition techniques for characterization of melt pool features.

In addition to predictive capabilities, expanding the workflow associated with our model implementation with a searchable database would reduce the time and cost involved in developing new high-performance materials for advanced manufacturing technologies. One of the practical motivations of our work is to understand and quantify characterization of melt pool geometries to enable their correlations to both manufacturing parameters and part quality. Toward that goal, we propose integration of the presented methodology with in-situ imaging data to guide real time control of additive manufacturing processes.

# Acknowledgments

This research was performed at the Department of Energy's Manufacturing Demonstration Facility located at Oak Ridge National Laboratory.

This manuscript has been authored by UT-Battelle, LLC, under contract DE-AC05-00OR22725 with the US Department of Energy (DOE). The US government retains and the publisher, by accepting the article for publication, acknowledges that the US government retains a nonexclusive, paid-up, irrevocable, worldwide license to publish or reproduce the published form of this manuscript, or allow others to do so, for US government purposes. DOE will provide public access to these results of federally sponsored research in accordance with the DOE Public Access Plan (http://energy.gov/downloads/doe-public-access-plan). Research was sponsored by the Advanced Manufacturing Office.

All authors contributed to research concept development and interpretation of the results. R. R. D. provided data set. V. C. P. formulated the project. A. P. designed and developed computational capability and wrote manuscript.

The code, trained models and data processing tools will be available upon release approval.